\documentclass[review]{elsarticle}

\usepackage{lineno}
\usepackage{amsmath}
\usepackage{graphicx}
\usepackage{amssymb}
\usepackage{amsfonts}
\usepackage{adjustbox}
\usepackage{pifont}
\usepackage{tabularx}
\usepackage{epsfig}
\usepackage{algorithmic}
\usepackage{balance}
\usepackage[lined,linesnumbered,ruled]{algorithm2e}
\usepackage{xcolor}
\usepackage[footnotesize]{caption}
\usepackage{multirow}
\usepackage{rotating}
\usepackage{lscape}
\usepackage{bm}
\usepackage{array}
\usepackage{url}
\usepackage {subfig,color}

\usepackage{amssymb} 
\usepackage{multirow}
\usepackage{booktabs}
\usepackage{diagbox}

\RequirePackage{latexsym,amsmath,amssymb}

\modulolinenumbers[5]

\journal{Journal of Elsevier}

\begin{document}

\begin{frontmatter}

\title{Attention-Guided Lidar Segmentation and Odometry Using Image-to-Point Cloud Saliency Transfer}

\author[mymainaddress,mysecondaryaddress]{Guanqun Ding}

\author[mysecondaryaddress,mythirdaddress]{Nevrez \.{I}mamo\u{g}lu\corref{mycorrespondingauthor}}
\ead{nevrez.imamoglu@aist.go.jp}
\cortext[mycorrespondingauthor]{Corresponding author.}
\author[mysecondaryaddress]{Ali Caglayan} 
\author[mymainaddress,mysecondaryaddress]{Masahiro Murakawa}
\author[mysecondaryaddress]{Ryosuke Nakamura}

\address[mymainaddress]{Graduate School of Science and Technology, University of Tsukuba, Tsukuba, 305-8577, Japan}
\address[mysecondaryaddress]{Digital Architecture Research Center (DigiARC), National Institute of Advanced Industrial Science and Technology, Tokyo, 135-0064, Japan}
\address[mythirdaddress]{CNRS-AIST Joint Robotics Laboratory (JRL), National Institute of Advanced Industrial Science and Technology, Tsukuba, 305-8560, Japan}

\begin{abstract}
LiDAR odometry estimation and 3D semantic segmentation are crucial for autonomous driving, which has achieved remarkable advances recently. However, these tasks are challenging due to the imbalance of points in different semantic categories for 3D semantic segmentation and the influence of dynamic objects for LiDAR odometry estimation, which increases the importance of using representative/salient landmarks as reference points for robust feature learning. To address these challenges, we propose a saliency-guided approach that leverages attention information to improve the performance of LiDAR odometry estimation and semantic segmentation models. Unlike in the image domain, only a few studies have addressed point cloud saliency information due to the lack of annotated training data. To alleviate this, we first present a universal framework to transfer saliency distribution knowledge from color images to point clouds, and use this to construct a pseudo-saliency dataset (i.e. FordSaliency) for point clouds. Then, we adopt point cloud-based backbones to learn saliency distribution from pseudo-saliency labels, which is followed by our proposed SalLiDAR module. SalLiDAR is a saliency-guided 3D semantic segmentation model that integrates saliency information to improve segmentation performance. Finally, we introduce SalLONet, a self-supervised saliency-guided LiDAR odometry network that uses the semantic and saliency predictions of SalLiDAR to achieve better odometry estimation. Our extensive experiments on benchmark datasets demonstrate that the proposed SalLiDAR and SalLONet models achieve state-of-the-art performance against existing methods, highlighting the effectiveness of image-to-LiDAR saliency knowledge transfer. Source code will be available at \emph{https://github.com/nevrez/SalLONet}.
\end{abstract}


\end{frontmatter}


\section{Introduction}
\label{sec_Introduction}

Understanding 3D point clouds has become increasingly important with the rise of robotics technologies such as augmented/virtual/mixed reality and autonomous vehicles. Autonomous driving, for instance, allows vehicles to sense and respond to their environment without human intervention. However, ensuring system safety relies heavily on accurate perception and localization of the environment. Simultaneous Localization and Mapping (SLAM) \cite{cadena2016past} technology plays a critical role in the perception and planning process of autonomous vehicles by constructing a map of the surrounding environment and localizing the vehicle. Visual/LiDAR odometry estimation \cite{wang2022approaches, zheng2021efficient, xu2021selfvoxelo, li2019net, wang2021pwclo} is an essential component of a SLAM system, aiming to estimate the robot's pose information from consecutive point clouds. Moreover, large-scale data-based applications, such as LiDAR semantic segmentation \cite{hou2022point, zhu2021cylindrical} and odometry estimation \cite{wang2021pwclo} empower advanced robotics technologies. Similarly, the use of saliency information in 2D computer vision tasks such as image translation \cite{jiang2021saliency}, object tracking \cite{liu2019aggregation, zhou2021saliency}, key-point selection \cite{tasse2015cluster, tinchev2021skd}, and person re-identification \cite{kim2021prototype, ren2022s, zhao2016person} has led to state-of-the-art results by capturing the pre-dominant information in a scene. However, the unstructured, unordered, and density-varied nature of point clouds makes it difficult for conventional point-cloud-based methods to process informative visual features effectively and rapidly in large-scale scenes. To address this challenge and enhance the performance of real- time autonomous vehicles, several works \cite{ding2019point, shtrom2013saliency, zheng2019pointcloud} have explored the use of saliency detection algorithms in point cloud data-based tasks, showing that the integration of efficient saliency knowledge can further enhance the performance of 3D point cloud understanding tasks. 

\begin{figure}[!ht]
    \centering
    \includegraphics[width=0.96\columnwidth]{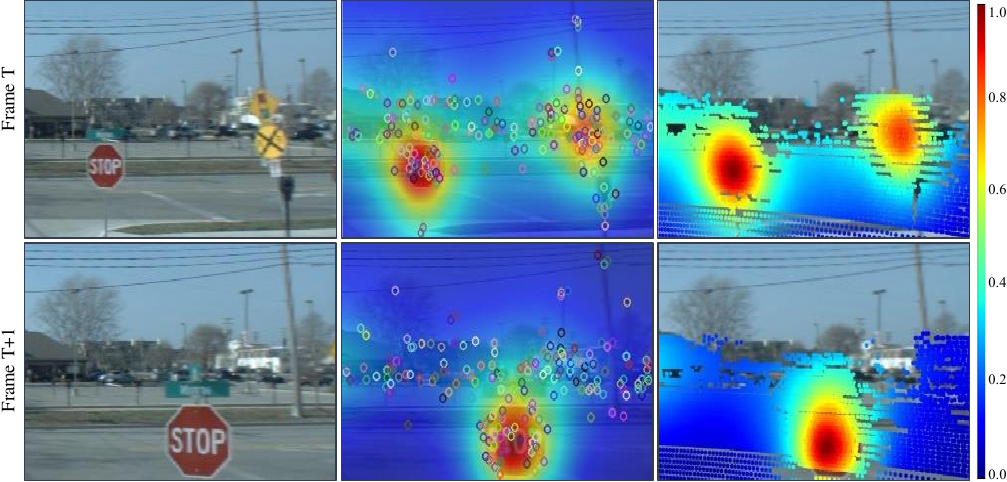}
    \caption{Visualization examples of SIFT \cite{lowe1999object} key points and saliency maps for two consecutive frames. From left to right column: RGB images, results of SIFT \cite{lowe1999object} key points and saliency maps from our FordSaliency dataset, and point clouds registered on images with saliency values.} 
\label{fig:keypoints_saliency}
\end{figure}

In LiDAR odometry estimation, keypoint selection is often used to facilitate the learning of matching features by the model \cite{zheng2020lodonet, liang2019salientdso, engel2017direct, prakhya2015sparse}. SIFT-based approaches, such as LodoNet \cite{zheng2020lodonet}, extract matched keypoint pairs, which are then used to learn point-wise features in Point-Net \cite{qi2017pointnet}. Alternatively, saliency-based point selection methods, such as that used by SalientDSO \cite{liang2019salientdso}, demonstrate the potential benefits of incorporating saliency information into visual odometry. As shown in Figure \ref{fig:keypoints_saliency}, while SIFT key points \cite{lowe1999object} and salient regions of saliency maps can both detect significant and consistent landmarks of the scene (e.g. buildings and traffic signs), saliency maps offer a more continuous and soft indication of attentive probabilities, unlike the sparse and discrete nature of key points. Thus, integrating saliency information has the potential to improve the performance of odometry estimation. However, despite significant progress in LiDAR odometry estimation, challenges remain, particularly in crowded environments with moving objects that can introduce noise and occlusions \cite{chen2019suma++,wang2021dsp, li2020dmlo, chen2021psf}. To address this issue, some odometry methods \cite{chen2019suma++,wang2021dsp, li2020dmlo, chen2021psf} use semantics to mitigate the adverse effects of moving object regions/points in the input data. Static objects can provide a stable and consistent reference point for geometry-based matching, which is critical for successful pose estimation. For example, early Iterative Closest Point (ICP) \cite{besl1992a, pomerleau2013comparing} based odometry models \cite{zhang2014loam, wang2021f} estimate the transformation iteratively by minimizing matching errors between corresponding points of two scans. 

\begin{figure}[!ht]
    \centering
    \includegraphics[width=0.92\columnwidth]{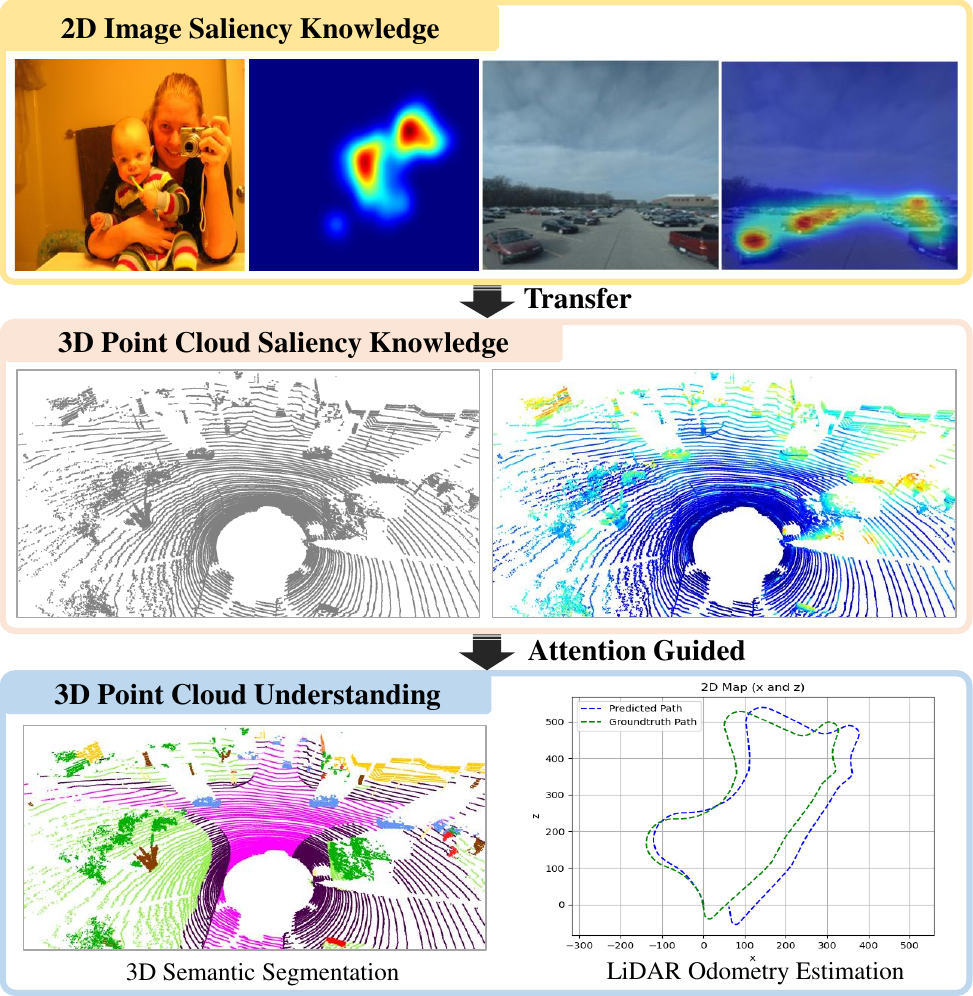}
  \caption{Overview of proposed framework of image-to-LiDAR saliency knowledge transfer for 3D point cloud understanding. The 2D images saliency knowledge of RGB saliency models is transferred to 3D point clouds. Then the 3D point cloud saliency knowledge is used to attention-guided 3D point cloud understanding tasks, such as 3D semantic segmentation and LiDAR odometry estimation.}
  \label{fig:framework_image_LiDAR}
\end{figure}

In this paper, we focus on improving LiDAR odometry estimation and 3D semantic segmentation by learning robust and discriminative features with saliency information constraints. Specifically, we propose a saliency-guided 3D semantic segmentation method that exploits saliency cues to facilitate the model in robust feature learning. Also, we propose a saliency-guided LiDAR odometry approach that leverages attention information and semantics to improve performance. Figure \ref{fig:framework_image_LiDAR} illustrates an overview of the proposed framework of image-to-LiDAR saliency knowledge transfer for attention-guided LiDAR semantic segmentation and odometry estimation models.

\begin{table*}
\caption{Comparison of existing saliency detection datasets on point clouds.
}
\centering
\begin{adjustbox}{width=\textwidth}
\begin{tabular}{ c | c | c | c | c | c | c | c | c }
\hline
\hline
\textbf{Dataset} & \textbf{Task} & \textbf{Pub.}& \textbf{\#Scan}& \textbf{\#Training}& \textbf{\#Testing} & \textbf{Type} &  \textbf{Annotation}& \textbf{Scene} \\
\hline
MeshSaliency \cite{chen2012schelling} &  Saliency Map Prediction &  TOG12 &  400 &  / &  / &  Mesh &  Key-points &  Objects. \\
\hline
PCSOD \cite{fan2022salient} &  Saliency Object Detection (SOD) &  ICCV22 &  2872 &  2000 &  872 &  Point Cloud &  Point-wise &  Indoor office, outdoor objects, etc. \\
\hline
\textbf{FordSaliency*} &  \textbf{Saliency Map Prediction} &  \textbf{Ours} &  \textbf{9920} &  \textbf{6103} &  \textbf{3817} &  \textbf{Point Cloud} &  \textbf{Point-wise} &  \textbf{Outdoor driving scenes.} \\
\hline
\hline
\multicolumn{9}{l}{* The saliency ground truth maps are created using the raw data of FordCampus dataset~\cite{pandey2011ford} through average response of saliency model-annotators (see \ref{subsub_fordsaliency}).} \\
\end{tabular}
\label{tab:pcsal_datasets_comp}
\end{adjustbox}
\end{table*}

Several attempts have been made to find effective solutions for saliency detection on point clouds \cite{ding2019point, shtrom2013saliency, tasse2015cluster, zheng2019pointcloud, ding2022sallidar}. In Table \ref{tab:pcsal_datasets_comp}, we compare the existing saliency detection datasets on point clouds. For saliency detection on point clouds, most challenges are yet to be explored further. First, previous saliency methods such as \cite{ding2019point, zheng2019pointcloud} have operated on mesh data of 3D objects, where scenes are less complicated with only a few background points. Second, due to the lack of human-annotated training datasets, it is unlikely to employ the supervised learning scheme for saliency detection on point clouds. Therefore, it is highly desired to develop a practicable pipeline based on deep learning for saliency prediction on large-scale point clouds. This study is an extension of our previous work \cite{ding2022sallidar} for point cloud saliency prediction and 3D semantic segmentation. In our previous work \cite{ding2022sallidar}, we have designed a universal framework and a point cloud saliency dataset (FordSaliency) to transfer saliency distribution knowledge for point clouds. Then an attention-guided two-stream network is proposed to improve the accuracy of LiDAR semantic segmentation task. The first stream is a LiDAR-based saliency network trained on FordSaliency dataset that guides the segmentation task. The second stream is a segmentation module that predicts the semantics of the input point cloud \cite{ding2022sallidar}. 

In this work, we propose a saliency-guided deep self-supervised odometry model that combines the saliency and semantic predictions of SalLiDAR [43] for the LiDAR odometry estimation.
In brief, the key and additional contributions of this work can be summarized as follows:
\begin{itemize}
    \item We propose a saliency-guided LiDAR odometry estimation model with a self-supervised learning manner. The proposed odometry model consists of three modules: saliency module, semantic module, and odometry module.
    \item To mitigate the adverse effects of dynamic points on the LiDAR odometry model, we binarize the semantic map into dynamic and static points using the semantic labels defined in the SemanticKITTI dataset [44]. The point cloud, along with the binarized semantic map and saliency map, is then fed into the odometry module for feature learning.
    \item To prioritize salient static points for point cloud matching, we propose a saliency-guided odometry loss that utilizes the saliency and binarized semantic maps to regulate the odometry module. This helps the module focus more on attentive points and improves the accuracy of point cloud matching.
Our extensive experiments on benchmark datasets suggest that the proposed two-stream LiDAR odometry model with saliency and semantic knowledge improves the performance of odometry estimation and achieves better performance compared with the existing methods.
\end{itemize}

\section{Related Works}
\label{sec_Related_Works}
\subsection{LiDAR Odometry Estimation} 
\label{subsec: lidar_odom}
Recently, deep learning-based odometry models \cite{nubert2021self, cho2020unsupervised, li2019net, wang2021pwclo} have been proposed to predict pose by learning more abundant features with powerful convolutional modules. 
In PWCLO-Net \cite{wang2021pwclo}, Wang \emph{et al.} propose a deep LiDAR odometry approach based on hierarchical
embedding mask optimization, where a warp-refinement module with attentive cost volume structure refines the estimated pose in a coarse-to-fine manner, the attentive cost volume is used for the association between two point clouds. In DeLORA \cite{nubert2021self} model, Nubert \emph{et al.} present a self-supervised LiDAR odometry model for pose regression without any ground-truth labels. Two consecutive range images converted from raw LiDAR point clouds are fed into DeLORA \cite{nubert2021self} model to output a rigid-body transformation, then the geometric transformation is applied to the source LiDAR scan and normal vector for obtaining transformed LiDAR scan and transformed normal vector. Afterward, a point-wise geometric loss between the transformed scan data and the target scan data can be calculated to guide the model to learn geometric-specific features, thus generating a transformation to match the transformed and target scans as closely as possible \cite{nubert2021self}. For this paper, we focus on the LiDAR odometry research based on deep learning, which has achieved great progress in recent years \cite{zheng2021efficient, li2019net, jonnavithula2021lidar, cho2020unsupervised}. 

\subsection{Saliency Detection on Point Cloud} 
\label{subsec: saliency_detection}

Saliency detection aims to find the most eye-attracting locations in a visual scene, which can be traced back to the pioneering work of Itti \emph{et al.} \cite{itti1998model}. With rapidly emerging advances and applications of deep learning techniques, saliency detection on color images/videos \cite{droste2020unified} has made great progress in recent years. There are also several works \cite{ding2019point, shtrom2013saliency, tasse2015cluster, tinchev2021skd, zheng2019pointcloud, chen2012schelling} for saliency computation on point clouds. 
For example, Ding \emph{et al.} \cite{ding2019point} propose a 3D mesh saliency calculation method by fusing local distinctness and global rarity features. 
Tinchev \emph{et al.} \cite{tinchev2021skd} present a key point detector on point clouds by using saliency estimation. They calculate the gradient response of a differentiable convolutional network to obtain the saliency map. Then they use multiple fully connected layers to combine the saliency feature, point cloud context feature, and PCA features of point descriptors \cite{tinchev2021skd}. Zheng \emph{et al.} \cite{zheng2019pointcloud} present a saliency computation method using a loss gradient approach that approximates point dropping in a differentiable manner of shifting points towards the point cloud center. 
However, saliency methods focusing on 3D meshes or indoor scenes are limited in their ability to process large-scale 3D point clouds such as 3D driving data. Also, saliency models extracting handcrafted descriptors may ignore informative representations for point clouds with varying density and complex backgrounds in outdoor scenarios.

\subsection{LiDAR Semantic Segmentation} 
\label{subsec: sem_seg}

LiDAR semantic segmentation \cite{hou2022point, hu2020randla, milioto2019rangenet++, zhu2021cylindrical, qi2017pointnet, xu2021rpvnet} is a crucial 3D computer vision task for autonomous driving, which aims to predict the semantic class of each point on a LiDAR scan. 
As a pioneering point set-based method, PointNet \cite{qi2017pointnet} uses multiple layer perceptrons (MLPs) to learn point-wise features for classification and segmentation. RandLA-Net \cite{hu2020randla} presents randomly sampling the input point cloud, and employs a local feature aggregation module to compensate for information loss introduced by the random sampling. 
Considering the range property of LiDAR point cloud, Cylinder3D \cite{zhu2021cylindrical} proposes a solution to leverage cylinder partition for 3D semantic segmentation. It also brings an asymmetrical model to encoder-decoder voxel-based features by 3D sparse convolutional networks. 
PVKD \cite{hou2022point} achieves the state-of-the-art performance of 3D semantic segmentation by applying the point-to-voxel knowledge distillation strategy to Cylinder3D \cite{zhu2021cylindrical} model. 
With RPVNet \cite{xu2021rpvnet}, the authors present a multi-modality fusion model that combines range-based, point-based, and voxel-based representations with a gated fusion module for LiDAR semantic segmentation.

\section{Proposed Framework}
\label{sec_Propose_Framework}
\subsection{Problem Formulation}
\label{Problem_Formulation}

Given an input point cloud $\mathcal{P}$=\{$p_{i} |$  $i$=$1,...,N$, $p_{i} \in \mathbb{R}^{d}$\} with a set of disordered points, where $N$ represents the point number of LiDAR frame and each point $p_{i}$ could contain $d$ dimensional features, such as point coordinates $(x, y, z)$, colors $(r, g, b)$, reflectivity, and normal feature. The objective of the saliency detection model on point cloud is to predict the saliency score map $\mathcal{S}$=\{$s_{i} |$  $i$=$1, ..., N$, $s_{i} \in [0, 1]$\}, where $s_{i}$ denotes the saliency score of point $p_{i}$. After normalizing the saliency prediction, the closer the saliency score $s_{i}$ to 1, the more attentive the point $p_{i}$. In the 3D semantic segmentation task, its goal is to predict the semantic class map $\mathcal{C}$=\{$c_{i} |$  $i$=$1, ..., N$, $c_{i} \in \mathbb{R}$\}, where $c_{i}$ indicates the semantic category of point $p_{i}$. 

The objective of this work is to establish a self-supervised LiDAR odometry estimation model that is guided by saliency and semantic constraints, and can be trained without ground-truth pose.
To achieve this goal, given the input of two consecutive LiDAR point clouds $\mathcal{P}_{t} \in \mathbb{R}^{d}$ and $\mathcal{P}_{t-1} \in \mathbb{R}^{d}$ at time $t$ and $t-1$ with a set of disordered points, where each point $p$ could contain $d$ dimensional point-wise features, such as point coordinates $(x, y, z)$, the range feature $r$, semantic feature $c$, and saliency feature $s$. The odometry model estimates a $3 \times 3$ rotational vector $\mathbf{q} \in SO(3)$ and a $3 \times 1$ translational vector $\mathbf{t}$, where the $\mathbf{R}$ and $\mathbf{t}$ compose the relative rigid transformation $\hat{T}_{t-1, t} \in SE(3)$ between point clouds $\mathcal{P}_{t}$ and $\mathcal{P}_{t-1}$. The $\mathcal{P}_{t}$ can be transformed into $\hat{\mathcal{P}}_{t-1}$ in the coordinate system of $\mathcal{P}_{t-1}$ by the transformation $\hat{T}_{t-1, t}$:
\begin{eqnarray}
    \hat{\mathcal{P}}_{t-1} = \hat{T}_{t-1, t} \odot \mathcal{P}_{t}
\label{equ:transform}
\end{eqnarray}
where $\odot$ represents the point-wise matrix multiplication. Afterward, the point-wise matching loss between $\hat{\mathcal{P}}_{t-1}$ and $\mathcal{P}_{t-1}$ can be calculated to train the odometry model, thereby forcing the model to predict an optimal transformation $\hat{T}_{t-1, t}$. Also, the normal vector $\mathcal{N}_{t}$ of $\mathcal{P}_{t}$ can be transformed into $\hat{\mathcal{N}}_{t-1}$ in the coordinate system of $\mathcal{P}_{t-1}$ by the transformation $\hat{T}_{t-1, t}$:
\begin{eqnarray}
    \hat{\mathcal{N}}_{t-1} = rot(\hat{T}_{t-1, t}) \odot \mathcal{N}_{t}
\label{equ:transform_normal}
\end{eqnarray}

Therefore, the odometry model can be trained in a self-supervised manner by calculating the point-wise matching loss, and it does not require the odometry ground truth $T_{t-1, t}$. 

\begin{figure*}[!ht]
    \centering
    \includegraphics[width=\textwidth]{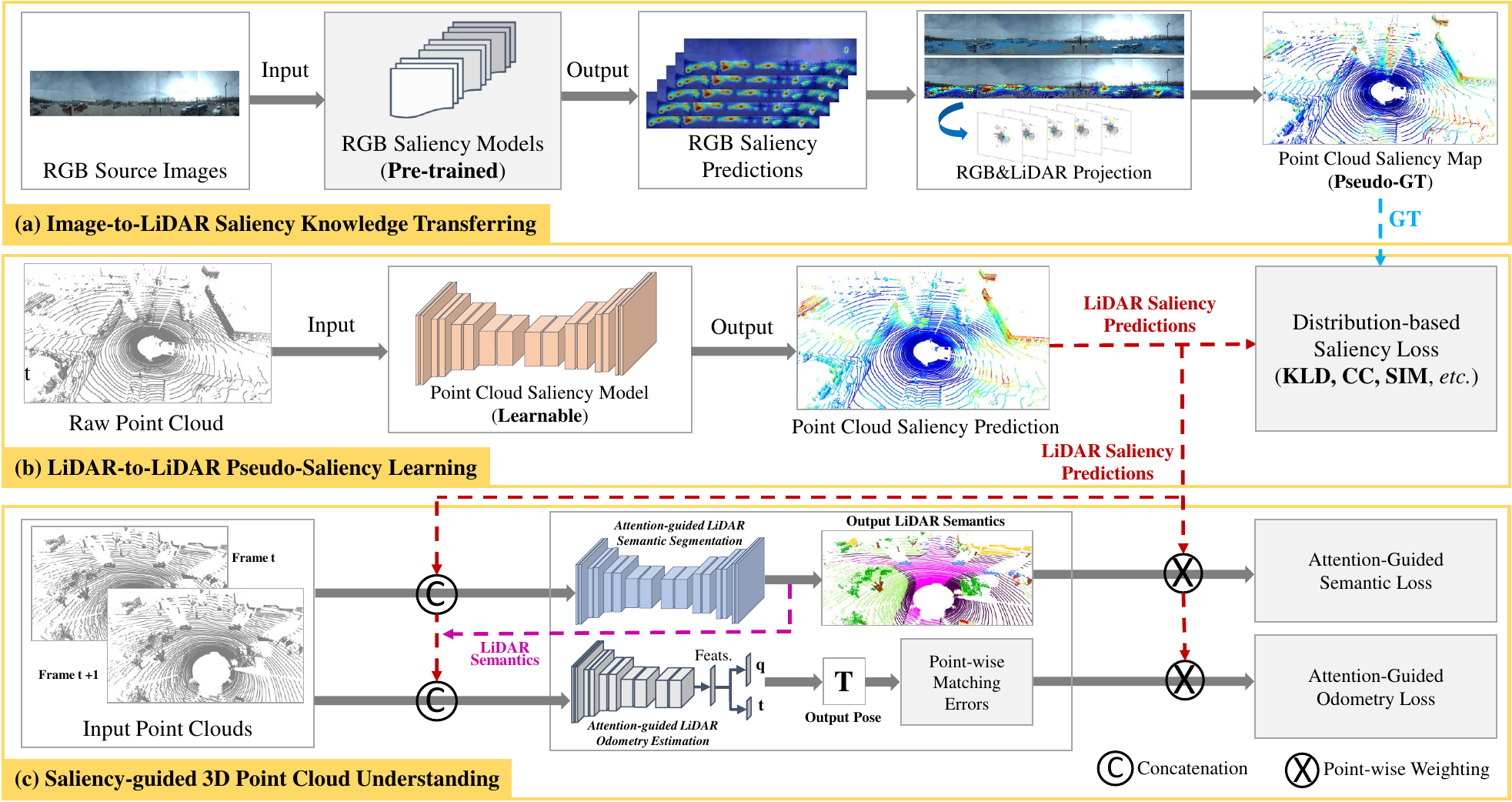}
  \caption{Illustration of proposed framework of image-to-LiDAR saliency knowledge transfer for attention-guided 3D point cloud understanding.}
  \label{fig:framework_overall}
\end{figure*}

\subsection{Framework Overview}
\label{overview}
In Figure \ref{fig:framework_overall}, we show the overview of image-to-LiDAR saliency knowledge transfer for 3D point cloud understanding. There are three main sub-tasks: 1) \emph{image-to-LiDAR saliency knowledge transferring} for generating a pseudo-saliency dataset of point clouds, 2) \emph{LiDAR-to-LiDAR pseudo-saliency learning} by using LiDAR-based deep models, and 3) \emph{saliency-guided 3D point cloud understanding} by integrating the saliency information.

\begin{figure}[!ht]
    \centering
    \includegraphics[width=0.9\columnwidth]{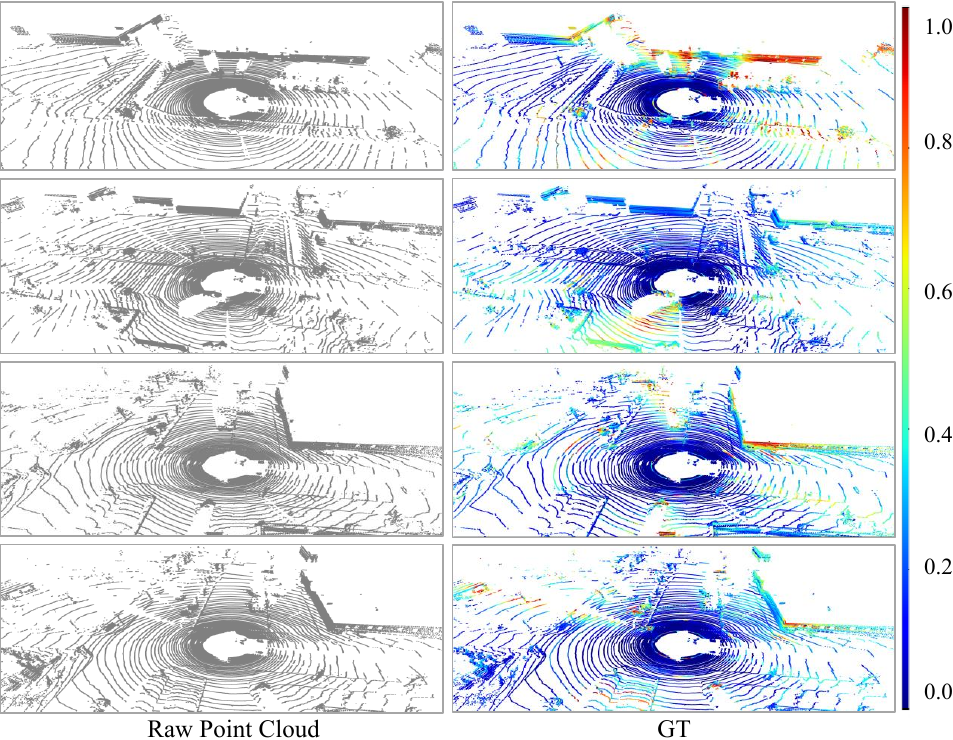}
  \caption{Visualization examples of point cloud and corresponding saliency pseudo-ground-truth (GT) map from our FordSaliency dataset.}
  \label{fig:saliency_fordsaliency}
\end{figure}

Firstly, we propose a large-scale pseudo-saliency dataset (FordSaliency) for point clouds by assigning the saliency values of RGB images to corresponding point clouds registered on images. In Figure~\ref{fig:saliency_fordsaliency}, we show the visualization examples of point cloud and corresponding saliency pseudo-ground-truth map from our FordSaliency dataset. Then, we train LiDAR-based models on the proposed pseudo-saliency dataset to learn point cloud saliency features. Next, we propose a saliency-guided two-stream network (\textbf{SalLiDAR}) for large-scale point cloud segmentation. 
The saliency prediction is not only used as an input feature for the semantic module but also adopted to saliency-guided loss to facilitate the semantic module 1) to learn more rich features of salient points and 2) to reduce the influence of the imbalance of points in different semantic classes with saliency constraints.

Finally, a \textbf{Sal}iency-guided \textbf{L}iDAR \textbf{O}dometry \textbf{Net}work (\textbf{SalLONet}) is proposed by applying the point cloud saliency and semantic information for performance improvement of odometry model. The objective of LiDAR odometry estimation is to output the pose information by matching two point clouds, in other words, it can be regarded as a problem of registration between two LiDAR scans. Therefore, our saliency-guided LiDAR odometry model is motivated by two observations: 1) the dynamic points should be suppressed as much as possible, since they may decrease
the performance of odometry estimation during the registration. On the other hand, 2) the static points should have more priority to make the model focus more on the salient static points for feature matching. To this end, we utilize our SalLiDAR model to predict semantics and the saliency map of the point cloud for odometry estimation. The semantic and saliency predictions are not only fed into the odometry model as input features but also integrated into a saliency-guided odometry loss to regularize the odometry model. 

\begin{figure*}[!ht]
\begin{center}
   \includegraphics[width=0.96\linewidth]{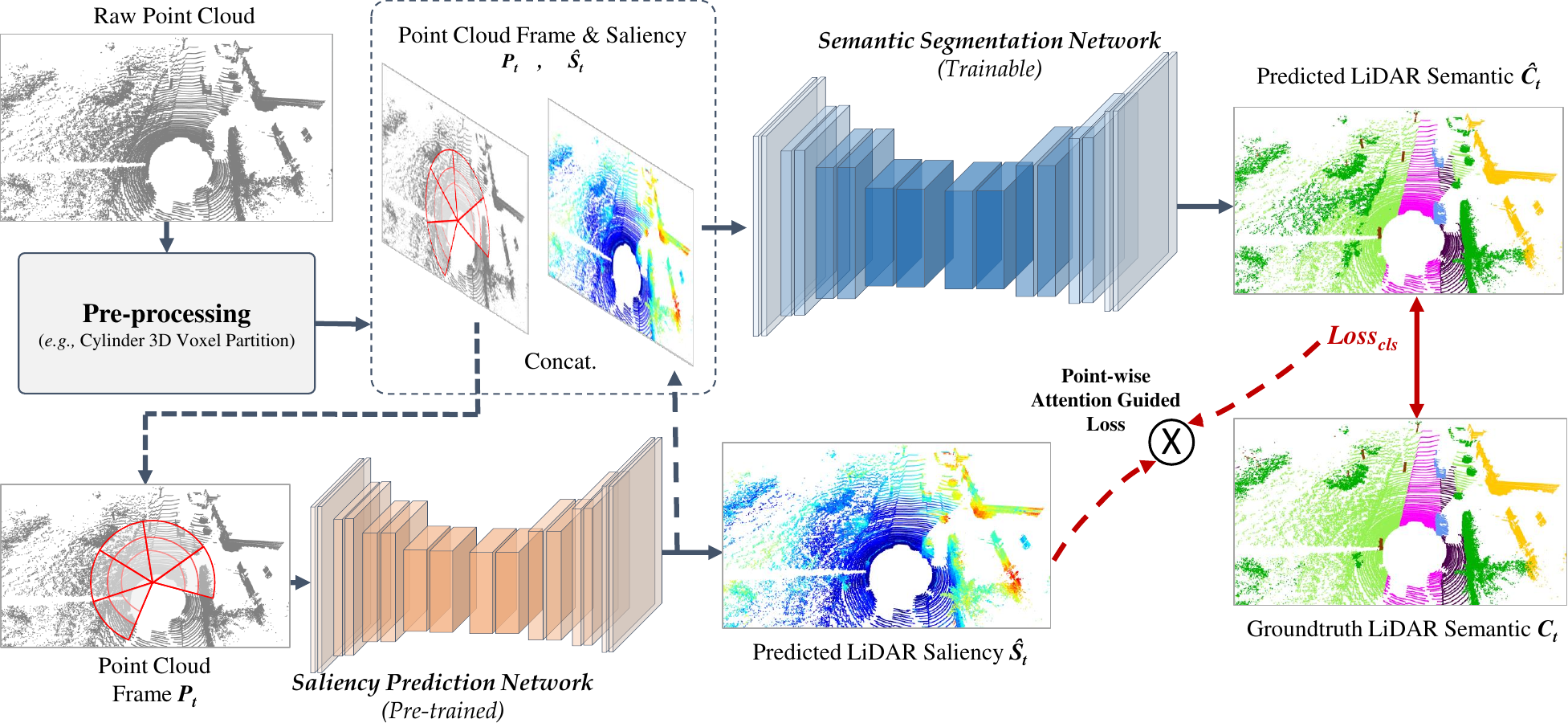} 
\end{center}
   \caption{Framework of proposed two-stream semantic segmentation model. The saliency prediction network is pre-trained on our FordSaliency dataset. 
}
\label{fig:sem_framework}
\end{figure*}

\subsection{Proposed Method}
\label{Proposed}
\noindent \textbf{\emph{1) Learning Point Cloud Saliency}}: 
In order to learn point cloud saliency representations, we adopt existing LiDAR-based semantic segmentation models \cite{hu2020randla, qi2017pointnet, zhu2021cylindrical} as backbones of the feature extractor. As shown in Figure~\ref{fig:framework_overall} (b), given a 3D LiDAR point cloud with coordinates and corresponding point-wise features, we first feed it into the feature extractor to obtain the representations of each point. Next, these learned features are passed by a saliency prediction layer to output the saliency score map of the input point cloud. 
We considered two types of models to learn saliency distribution on point clouds: i) classification-based saliency prediction and ii) commonly used saliency regression. More details can be referred to our previous work \cite{ding2022sallidar}.

\noindent \textbf{\emph{2) Two-Stream Segmentation Model}}: 
As depicted in Figure~\ref{fig:sem_framework}, we develop a two-stream semantic segmentation model on the point cloud by combining features from the saliency module and the semantic module. We feed an input point cloud into the saliency branch to predict the saliency distribution of the whole scene. Meanwhile, the point cloud is also fed into the semantic branch to extract point features and output the predictions of the semantic class. To validate the effectiveness of the learned point cloud saliency distribution knowledge, we initialize and freeze the parameters of the saliency branch with the weights pre-trained on the FordSaliency dataset. More details can be referred to our previous work \cite{ding2022sallidar}.



\begin{figure*}[!ht]
    \centering
    \includegraphics[width=0.98\textwidth]{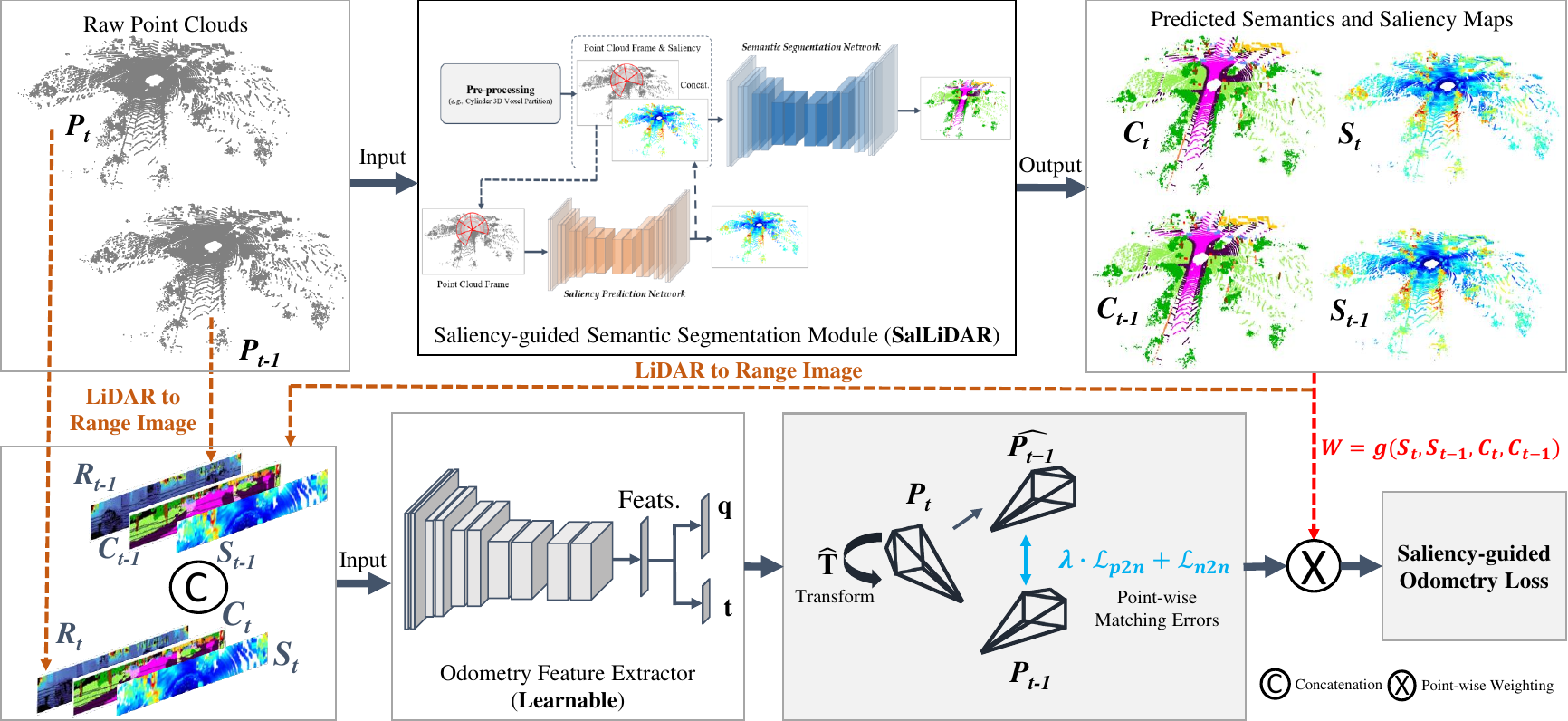}
    \caption{Architecture of \textbf{Sal}iency-guided \textbf{L}iDAR \textbf{O}dometry \textbf{Net}work (\textbf{SalLONet}). The \textbf{\emph{$\mathcal{C}_{t}$}}, \textbf{\emph{$\mathcal{C}_{t-1}$}}, \textbf{\emph{$\mathcal{S}_{t}$}}, \textbf{\emph{$\mathcal{S}_{t-1}$}}, \textbf{\emph{$\mathcal{R}_{t}$}}, \textbf{\emph{$\mathcal{R}_{t-1}$}} represent the predicted LiDAR semantic maps, LiDAR saliency maps, and range images of corresponding two consecutive point clouds \textbf{\emph{$\mathcal{P}_{t}$}} and \textbf{\emph{$\mathcal{P}_{t-1}$}}. The transformation \textbf{$\Hat{T}$} of the two-point clouds consists of the estimated translation \textbf{t} and rotation \textbf{q} (i.e., pose).}
\label{fig:sallonet_framework}
\end{figure*}

\noindent \textbf{\emph{3) LiDAR Odometry Estimation Module}}: 
As shown in Figure \ref{fig:sallonet_framework}, the semantic map and saliency map are first predicted by the proposed saliency and semantic modules of SalLiDAR. For odometry estimation, we convert and concatenate two consecutive LiDAR point clouds and their respective predicted saliency and semantic maps to range images as the input of the odometry module. The outputs of the odometry module are the feature vectors of translation $\textbf{t}$ and rotation $\textbf{q}$ between two LiDAR point clouds. Then we can construct the rigid body transformation $\hat{T}_{t-1, t}$ based on the predicted translation and rotation. The source LiDAR scan $\mathcal{P}_{t}$ can be transformed into $\hat{\mathcal{P}}_{t-1}$ by the transformation $\hat{T}_{t-1, t}$ for matching with the target LiDAR scan $\mathcal{P}_{t-1}$. Thus, the odometry module can be supervised by the point-wise matching errors between the transformed scan $\Hat{\mathcal{P}}_{t-1}$ and the target scan $\mathcal{P}_{t-1}$. 

\begin{figure}[!ht]
    \centering
    \includegraphics[width=0.78\columnwidth]{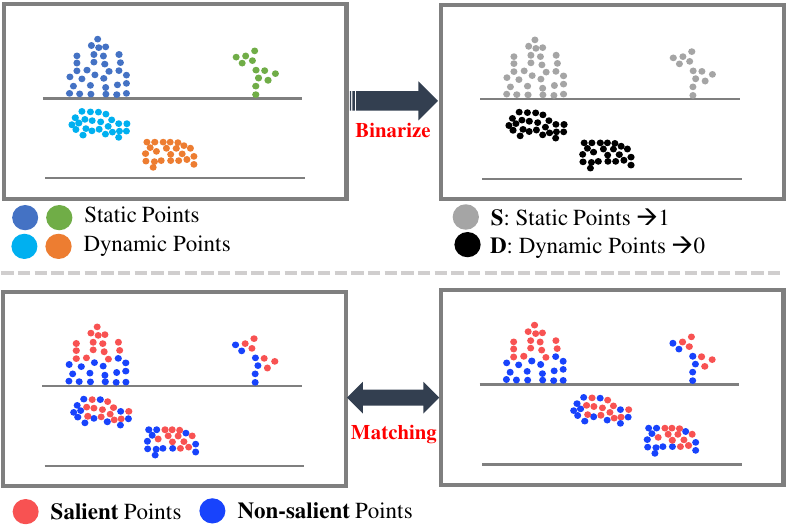}
    \caption{Demonstrations of semantics and saliency map for LiDAR odometry estimation. The top depicts the binarized semantics based on dynamic (e.g., car) and static point (e.g., building, traffic sign), and the bottom shows the point-wise matching with saliency maps between two consecutive scans.}
\label{fig:sallonet_bin_dynamic_points}
\end{figure}

To regularize the odometry module, we exploit the saliency and semantic segmentation information to saliency-guided odometry loss for the model training. The previous odometry studies \cite{li2019net, wang2021pwclo} have shown that dynamic points may degrade the performance of odometry estimation, thus we exploit the predicted semantic map to suppress the effect of dynamic points. On the other hand, we utilize the predicted saliency map to increase the priorities of static salient points for matching two LiDAR scans. To obtain the saliency and semantic predictions for the LiDAR odometry module, we initialize and freeze the parameters of the saliency branch with the weights pre-trained on our FordSaliency \cite{ding2022sallidar} dataset. We also use the weights learned from the SemanticKITTI \cite{behley2019semantickitti} dataset to initialize and freeze the parameters of the semantic module. Since the predicted saliency distribution represents the attention level of each point, we can apply it to constrain the parameter optimization of the semantic segmentation module. Similar to the proposed SalLiDAR model, we propose three different integration methods with saliency semantic information for LiDAR odometry estimation:

\begin{table}
\caption{Assignment of dynamic (D$\rightarrow$0) and static (S$\rightarrow$1) points based on semantic categories defined in
SemanticKITTI \cite{behley2019semantickitti} dataset.
}
\centering
\begin{adjustbox}{width=0.48\textwidth}
\begin{tabular}{ c | c }
\hline
\hline
\textbf{Dynamic Class (D$\rightarrow$0)} & \textbf{Static Class (S$\rightarrow$1)} \\
\hline
{car} &  {road} \\
{bicycle} & {parking} \\
{motorcycle} & {sidewalk} \\
{truck} & {other-ground} \\
{other-vehicle} & {building} \\
{person} & {fence} \\
{bicyclist} & {vegetation} \\
{motorcyclist} & {trunk} \\
~  & {terrain} \\
~  & {pole} \\
~  & {traffic-sign} \\
\hline
\hline
\end{tabular}
\label{tab:bin_dynamic_static_sem}
\end{adjustbox}
\end{table}

\noindent \textbf{SalLONet-I: Saliency-guided odometry loss with saliency prediction and semantic mask for odometry estimation.} If we remove saliency and semantic concatenation module in Figure \ref{fig:sallonet_framework}, it becomes SalLONet-I. Considering that the dynamic points (e.g., moving car, pedestrian) may introduce adverse results on the odometry estimation, we first convert the predicted semantic map to a binarized mask to indicate the static points (e.g., building, road) and dynamic points (e.g., car, person). As shown in Figure \ref{fig:sallonet_bin_dynamic_points} (top) and Table \ref{tab:bin_dynamic_static_sem}, the predicted semantics are binarized to dynamic and static points based on the semantic categories defined in SemanticKITTI \cite{behley2019semantickitti} dataset. The point with a semantic class of moving or potentially moving is defined as a dynamic point. For example, regardless of whether the semantic category of a point is a \emph{car} or a \emph{moving car}, it will be defined as a dynamic point. Then, the binarized semantic mask is applied to suppress the adverse effects of the dynamic points and increase the weights of static points for the point-wise matching odometry loss. Additionally, we apply saliency prediction to odometry loss for odometry model training, thus facilitating the odometry model to focus more on the static salient points for matching, as shown in Figure \ref{fig:sallonet_bin_dynamic_points} (bottom). 

By following the study \cite{nubert2021self}, we use the geometric-based losses to optimize the odometry estimation module by calculating the point-wise matching errors:
\begin{eqnarray}
    \mathcal{L}^{odom} = \lambda \cdot \mathcal{L}_{p2n} + \mathcal{L}_{n2n} 
\label{equ:odom_loss_geometric}
\end{eqnarray}
where $\lambda$ is a constant to balance the two losses. $\mathcal{L}_{p2n}$ and $\mathcal{L}_{n2n}$ are point-to-plane loss and point-to-point loss, which can be represented as follows: 
\begin{eqnarray}
    \mathcal{L}_{p2n} = \frac{1}{N} \sum_{i=1}^{N} | (\Hat{p}_{i} - {p}_{i}) \cdot {n}_{i} |_{2}^{2} \\
    \mathcal{L}_{n2n} = \frac{1}{N} \sum_{i=1}^{N} | \Hat{n}_{i} - {n}_{i} |_{2}^{2}
\label{equ:odom_loss_geometric2}
\end{eqnarray}
where $\Hat{p}_{i}$ and ${p}_{i}$ are the point coordinate values of $\Hat{P}$ and ${P}$, respectively. $\Hat{n}_{i}$ and ${n}_{i}$ are the normal values of $\Hat{\mathcal{N}}$ and $\mathcal{N}$, respectively.

\begin{figure}[!ht]
    \centering
    \includegraphics[width=0.9\columnwidth]{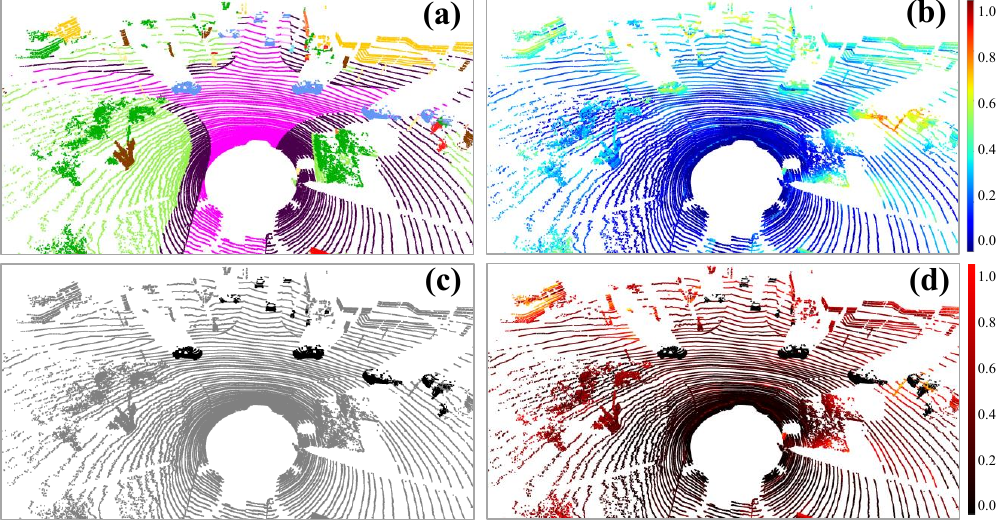}
    \caption{Visualization of saliency and semantic maps for odometry estimation. The results include (a) semantic prediction and (b) saliency prediction of the proposed SalLiDAR model; (c) binarized semantic map (i.e. dynamic and static points); and (d) weighted map by saliency and binarized semantics for the loss weighting of proposed SalLONet by Eq. \ref{equ:weighted_odom_loss_sallonet}.}
\label{fig:visualize_salsem_loss}
\end{figure}

To guide the odometry model to focus more on salient static points for matching, we apply the saliency and binarized semantic maps to the geometric-based loss. Thus, the saliency-guided odometry loss can be represented as: 
\begin{eqnarray}
    \hat{\mathcal{L}}^{odom} = \frac{1}{N} \sum_{i=1}^{N} {l}^{odom}_{i} * \mathcal{W} \\
    \mathcal{W} = \exp{(s_{i}^{t}*{s}_{i}^{t-1}*{c}_{i}^{t}*{c}_{i}^{t-1})}
\label{equ:weighted_odom_loss_sallonet}
\end{eqnarray}
where $i$ is the index of point; $\hat{\mathcal{L}}^{odom}$ is the weighted loss for odometry; ${l}^{odom}_{i}$ denotes the point-wise odometry matching loss of point $p_{i}$. $s_{i}^{t}$ and ${s}_{i}^{t-1}$ denote the saliency prediction of point $p_{i}$ from scan $t$ and scan $t-1$; $c_{i}^{t}$ and ${c}_{i}^{t-1}$ denote the binarized semantic predictions of point $p_{i}$ from scan $t$ and scan $t-1$; the $*$ represents the element-wise multiplication. Figure \ref{fig:visualize_salsem_loss} illustrates a visualization of the saliency and semantic maps used for attention-guided LiDAR odometry estimation. We can observe that the weights of dynamic points (e.g., car) are suppressed, while the static points (e.g., building) are highlighted by the binarized semantics and LiDAR saliency map.

\noindent \textbf{SalLONet-II: Saliency prediction and semantic mask as descriptors for odometry estimation.} If we remove point-wise saliency guided loss module in Figure \ref{fig:sallonet_framework}, it becomes SalLONet-II. We append the normalized saliency and binarized semantic prediction to point cloud coordinates as input features for the odometry model. We believe that prior saliency knowledge and high-level semantic information could be helpful for feature learning and localization in odometry estimation. 

\noindent \textbf{SalLONet-III: Saliency distribution and semantic prediction as descriptors and attentive loss guiding for odometry estimation.} Figure \ref{fig:sallonet_framework} shows the SalLONet-III model. In this model, the saliency and semantic maps are not only utilized as the additional input features for the odometry module but also applied to saliency-guided odometry loss for optimization during training.

\section{Experimental Analysis}
\label{sec_Experimental_Analysis}

\subsection{\textbf{Implementation Details}} 
\label{sub_imple}

For point cloud saliency detection, we employ PointNet \cite{qi2017pointnet}, RandLA-Net \cite{hu2020randla}, and Cylinder3D \cite{zhu2021cylindrical} models as feature extractors. 
For 3D semantic segmentation, we use RandLA-Net \cite{hu2020randla} and Cylinder3D \cite{zhu2021cylindrical} as baselines.
For LiDAR odometry, we adopt the DeLORA \cite{nubert2021self} model as the baseline of LiDAR odometry estimation. By following follow DeLORA \cite{nubert2021self} model, the range images and normal features are converted from two consecutive raw LiDAR point clouds as input features for odometry estimation. We adopt our SalLiDAR model \cite{ding2022sallidar} to generate saliency and semantic predictions for the LiDAR odometry module. For a fair comparison, all the odometry networks of baseline and our proposed methods are randomly initialized, and they are trained on KITTI \cite{geiger2012we} odometry dataset from scratch with 5-fold validation. The initial learning rate is 1e-5, and all odometry models are trained with self-supervised learning for 100 epochs.

\subsection{\textbf{Datasets and Experimental Setup}} 
\label{sub_datasets}

\subsubsection{\textbf{LiDAR FordSaliency Dataset}} 
\label{subsub_fordsaliency}

Based on the data of the FordCampus \cite{pandey2011ford} dataset, we build a point cloud saliency dataset (namely FordSaliency) for the training of LiDAR-based saliency models. 
We utilize dataset1 and dataset2 of FordSaliency as the validation set and training set, respectively. More details can be referred to the work \cite{ding2022sallidar}.

\subsubsection{\textbf{SemanticKITTI Dataset}} 
\label{subsub_SemanticKITTI}

SemanticKITTI \cite{behley2019semantickitti} dataset is a well-known large-scale dataset for point cloud semantic segmentation. This dataset consists of 22 Velodyne driving-scene sequences, which are split into a training set (sequences 00-07 and 09-10), a validation set (sequence 08), and a testing set (sequences 11-21). 

\subsubsection{\textbf{KITTI Odometry Dataset}} 
\label{subsub_KITTIOdom}

We conduct all the odometry experiments on KITTI \cite{geiger2012we} odometry dataset, which provides LiDAR point clouds captured from the Velodyne lidar sensor.
By following the odometry works \cite{nubert2021self, li2019net}, the dataset is divided into a training set (Sequences 00-08) and a validation set (Sequences 09-10).

\subsection{\textbf{Evaluation Metrics}} 
\label{sub_evalmetric}

For point cloud saliency detection, we use popular saliency metrics\footnote{\url{https://saliency.tuebingen.ai/evaluation.html}} including Correlation Coefficient (CC), Similarity (SIM), and Kullback-Leibler Divergence (KLD) to evaluate the performance of point cloud saliency model. For performance evaluation of LiDAR semantic segmentation, we adopt mean Intersection-over-Union (mIoU)\footnote{http://semantic-kitti.org/tasks.html} as evaluation metric following the previous studies \cite{hu2020randla, zhu2021cylindrical}.
For LiDAR odometry estimation, the average translational ($[\%]$) and average rotational ($[\frac{deg}{100m}]$) RMSE (Root Mean Square Error)\footnote{https://www.cvlibs.net/datasets/kitti/eval\_odometry.php} are adopted to evaluate the performance of LiDAR odometry models.

\subsection{\textbf{Results and Performance Analysis}} 
\label{sub_results}

\subsubsection{LiDAR Saliency Results on FordSaliency Dataset}
\label{exp_results_saliency}
\begin{figure}[!ht]
\begin{center}
   \includegraphics[width=\linewidth]{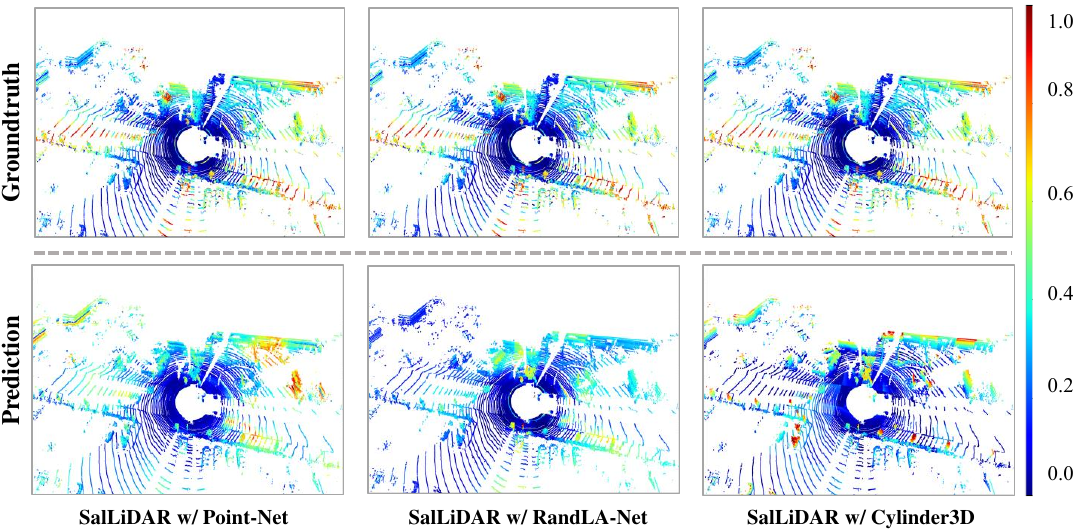} 
\end{center}
   \caption{Point cloud saliency prediction results of SalLiDAR model with different backbones on FordSaliency dataset. 
   }
\label{fig:sal_predall}
\end{figure}

We compare the performance of LiDAR-based saliency models with different feature extractors on our FordSaliency dataset. 
In Figure~\ref{fig:sal_predall}, we show the visualization results of SalLiDAR models with different backbones on the FordSaliency validation set. 
In Table~\ref{tab_saliency_val}, we report the quantitative performance of these models on the FordSaliency validation set. From 
Figure~\ref{fig:sal_predall} and
Table~\ref{tab_saliency_val}, we can observe that although the saliency annotations are pseudo-labels, all these LiDAR-based models are able to learn the discriminative point cloud saliency representations for saliency distribution prediction. 
On the other hand, the model with the Cylinder3D backbone can predict better saliency distribution than the model with other backbones. The models with RandLA-Net backbone and PointNet backbone can learn the correlation and similarity features from point cloud saliency annotations, as evidenced by the CC, SIM, and KLD values in Table~\ref{tab_saliency_val}. However, the prediction of the model with the Cylinder3D backbone can achieve higher CC, SIM, and lower KLD performance. It suggests that the model with a voxel-based partition (\emph{e.g.} 3D Cylinder) could learn more powerful saliency representations than point-based models.

\begin{table}[t]
\caption{Results of SalLiDAR models with different backbones on FordSaliency dataset. 
Note that the larger the values of CC, SIM, and the smaller the value of KLD, the better the performance of the point cloud saliency approach.
}
\centering
\begin{adjustbox}{width=0.7\columnwidth}
\begin{tabular}{c |c c c }
\hline  
\hline
\textbf{Saliency Model}  & \textbf{CC $\uparrow$} & \textbf{SIM$\uparrow$} &  \textbf{KLD$\downarrow$}  \\
\hline
SalLiDAR w/ PointNet & 0.3465 & 0.6655 & 0.4263  \\
SalLiDAR w/ RandLA-Net & 0.6368 & 0.7784 & 0.1688  \\
SalLiDAR w/ Cylinder3D & \textbf{0.6760} & \textbf{0.7854} & \textbf{0.1574} \\
\hline
\hline
\end{tabular}
\end{adjustbox}
\label{tab_saliency_val}
\end{table}

\begin{figure*}[!ht]
\begin{center}
   \includegraphics[width=\linewidth]{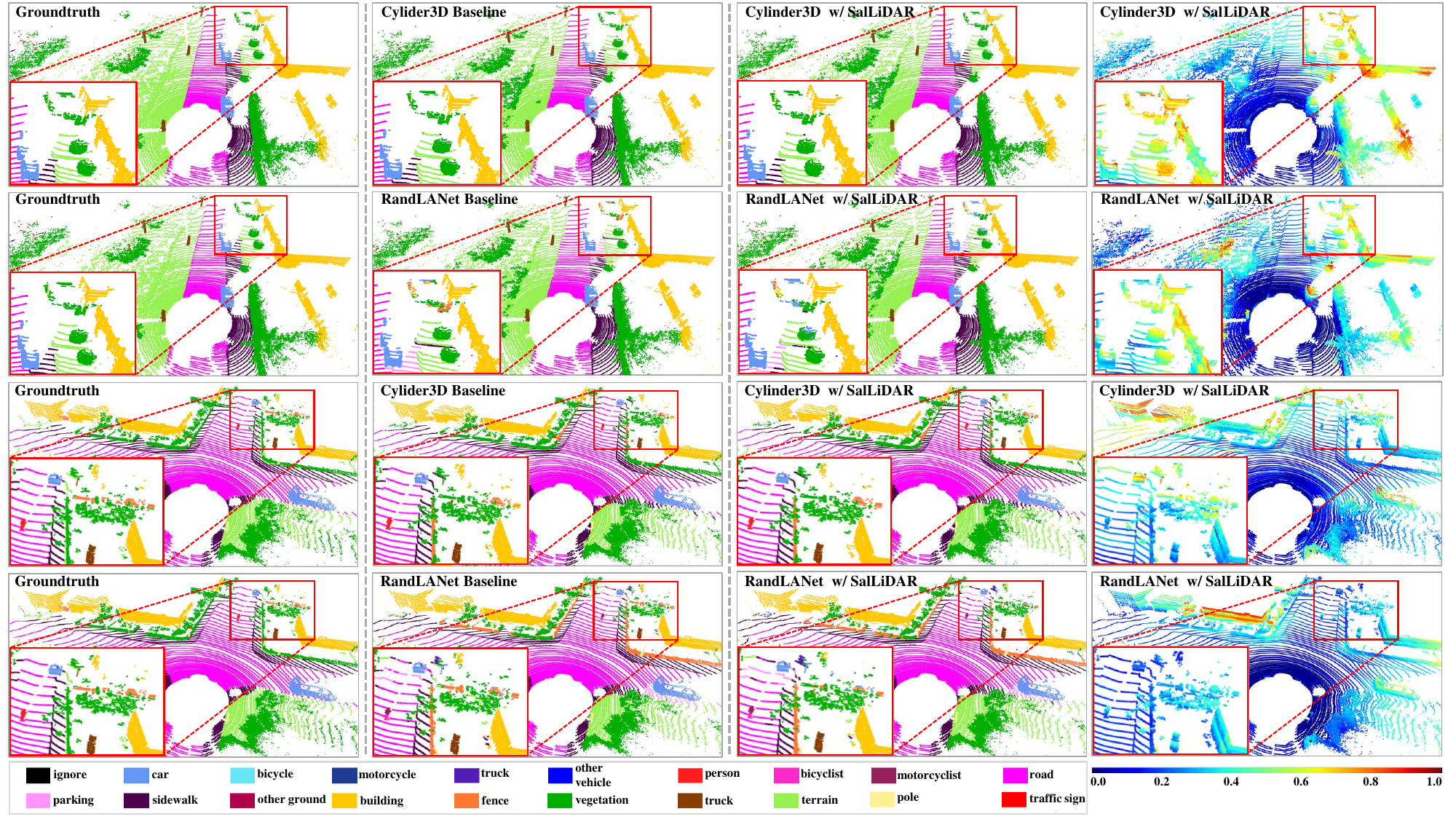} 
\end{center}
   \caption{Visualization comparison results of baseline and proposed LiDAR segmentation models on SemanticKITTI \cite{behley2019semantickitti}. From the first column to the last column are: the visualizations of semantic ground-truth, the semantic predictions of baseline models, the semantic results of proposed models, and the saliency predictions of proposed models, respectively.
}
\label{fig:sem_pred_comp}
\end{figure*}

\begin{table*}[!htp]
\caption{Performance comparison of proposed models and existing LiDAR segmentation methods on SemanticKITTI \cite{behley2019semantickitti} test set. Results are obtained from the leaderboard and literature.}
\label{semantickitti}
\centering
\begin{adjustbox}{width=\textwidth}
\begin{tabular}{c c c c c c c c c c c c c c c c c c c c c}
\hline
\textbf{Methods} & \textbf{mIoU} & \rotatebox{90}{car} &  \rotatebox{90}{bicycle} & \rotatebox{90}{motorcycle} & \rotatebox{90}{truck} & \rotatebox{90}{other-vehicle} & \rotatebox{90}{person} & \rotatebox{90}{bicyclist} & \rotatebox{90}{motorcyclist} & \rotatebox{90}{road} & \rotatebox{90}{parking} & \rotatebox{90}{sidewalk} & \rotatebox{90}{other-ground} &
\rotatebox{90}{building} & \rotatebox{90}{fence} & \rotatebox{90}{vegetation} & \rotatebox{90}{trunk} & \rotatebox{90}{terrain} & \rotatebox{90}{pole} & \rotatebox{90}{traffic-sign} \\
\hline
\hline
Darknet53~\cite{behley2019semantickitti} & 49.9 & 86.4 & 24.5 & 32.7 & 25.5 & 22.6 & 36.2 & 33.6 & 4.7 & \textcolor{red}{91.8} & 64.8 & 74.6 & {27.9} & 84.1 & 55.0 & 78.3 & 50.1 & 64.0 & 38.9 & 52.2 \\
RangeNet++~\cite{milioto2019rangenet++} & 52.2 & 91.4 & 25.7 & 34.4 & 25.7 & 23.0 & 38.3 &  38.8 & 4.8 & \textcolor{red}{91.8} & {65.0} & 75.2 & 27.8 & 87.4 & 58.6 & 80.5 & 55.1 & 64.6 & 47.9 & 55.9 \\
RandLA-Net~\cite{hu2020randla} & 53.9 & 94.2 & 26.0 & 25.8 & 40.1 & 38.9 & 49.2 & 48.2 & 7.2  & 90.7 & 60.3 & 73.7 & 20.4 & 86.9 & 56.3 & 81.4 & 61.3 & 66.8 & 49.2 & 47.7 \\ 
PolarNet~\cite{zhang2020polarnet} & 54.3 & 93.8 & 40.3 & 30.1 & 22.9 & 28.5 & 43.2 & 40.2 & 5.6 & 90.8 & 61.7 & 74.4 & 21.7 & {90.0} & 61.3 & 84.0 & 65.5 & 67.8 & 51.8 & 57.5  \\
SqueezeSegv3~\cite{xu2020squeezesegv3} & 55.9 & 92.5 & 38.7 & 36.5 & 29.6 & 33.0 & 45.6 & 46.2 & {20.1} & 91.7 & 63.4 & 74.8 & 26.4 & 89.0 & 59.4 & 82.0 & 58.7 & 65.4 & 49.6 & 58.9  \\
Salsanext~\cite{cortinhal2020salsanext} & 59.5 & 91.9 & 48.3 & 38.6 & 38.9 & 31.9 & 60.2 & 59.0 & 19.4 & 91.7 & 63.7 & 75.8 & 29.1 & 90.2 & 64.2 & 81.8 & 63.6 & 66.5 & 54.3 & 62.1 \\
KPConv~\cite{thomas2019kpconv} &58.8& 96.0&32.0 & 42.5 & 33.4&44.3&61.5 & 61.6 & 11.8 & 88.8 & 61.3&  72.7&31.6& \textcolor{red}{95.0} & 64.2 & 84.8 & 69.2 & 69.1 & 56.4 & 47.4 \\
FusionNet~\cite{zhang2020deep} & 61.3 & 95.3 & 47.5 & 37.7 & 41.8 & 34.5 & 59.5 & 56.8 & 11.9 & \textcolor{red}{91.8} & 68.8 & 77.1 & 30.8 & 92.5 & {69.4} & 84.5 & 69.8 & 68.5&60.4 & {66.5} \\ 
KPRNet~\cite{kochanov2020kprnet} & 63.1 & 95.5&54.1& 47.9&23.6 & 42.6&65.9 & 65.0 & 16.5 & {93.2} & {73.9} & \textcolor{red}{80.6} & 30.2 & 91.7 & {68.4} & {85.7} & 69.8 & 71.2 & 58.7 & 64.1 \\
TORANDONet~\cite{gerdzhev2021tornado} & 63.1 &94.2& 55.7& 48.1& 40.0& 38.2& 63.6& 60.1& 34.9& 89.7& 66.3& 74.5& 28.7& 91.3& 65.6& 85.6& 67.0& {71.5} & 58.0 & {65.9} \\
SPVNAS~\cite{tang2020searching} & 66.4 & \textcolor{red}{97.3} & 51.5 & 50.8 & 59.8 & 58.8 & 65.7 & 65.2 & 43.7 & 90.2 & 67.6 & 75.2 & 16.9 & 91.3 & 65.9 & 86.1 & 73.4 & 71.0 & 64.2 & \textcolor{red}{66.9} \\
Cylinder3D~\cite{zhu2021cylindrical} & 67.8 & 97.1 & 67.6 & 64.0 & 59.0 & 58.6 & 73.9 & 67.9 & 36.0 & {91.4} & {65.1} & {75.5} & 32.3 & {91.0} & {66.5} & {85.4} & 71.8 & {68.5} & 62.6 & {65.6}  \\
PVKD \cite{hou2022point} & 71.2 & 97.0 & 67.9 & {69.3} & 53.5 & {60.2} & {75.1} & {73.5} & \textcolor{red}{50.5} & \textcolor{red}{91.8} & \textcolor{red}{70.9} & 77.5 & {41.0} & 92.4 & {69.4} & {86.5} & \textcolor{red}{73.8} & \textcolor{red}{71.9} & {64.9} & 65.8 \\
\hline
* RandLA-Net (baseline) & 52.5 & 93.8 & 27.0 & 22.0 & 36.1 & 38.1 & 49.9 & 42.5 & 6.4  & 90.7 & 58.8 & 74.1 & 11.5 & 88.9 & 57.4 & 79.8 & 61.2 & 65.5 & 49.9 & 46.0 \\ 
\textbf{RandLA-Net+SalLiDAR-I} & \textbf{54.0} & \textbf{94.1} & \textbf{28.2} & \textbf{24.4} & \textbf{45.4} & 37.2 & 48.3 & \textbf{48.1} & 5.9 & 89.1 & \textbf{59.7} & 72.4 & \textbf{21.9} & 87.5 & 56.2 & \textbf{81.7} & \textbf{61.6} & \textbf{68.6} & 49.7 & \textbf{46.5}  \\
\textbf{RandLA-Net+SalLiDAR-II} & \textbf{53.4} & 93.8 & \textbf{30.2} & \textbf{24.3} & \textbf{37.9} & 37.5 & \textbf{50.1} & \textbf{45.5} & \textbf{9.5} & 89.9 & \textbf{60.1} & 73.9 & \textbf{13.8} & 87.3 & 56.6 & \textbf{81.3} & 60.7 & \textbf{67.2} & 48.0 & \textbf{47.8}  \\
\textbf{RandLA-Net+SalLiDAR-III} & \textbf{53.8} & \textbf{94.4} & \textbf{28.9} & \textbf{26.6} & 35.5 & \textbf{39.7} & 47.0 & \textbf{47.2} & \textbf{11.3} & 90.0 & \textbf{60.5} & 73.7 & \textbf{16.2} & 88.3 & 56.8 & \textbf{81.3} & 60.9 & \textbf{67.8} & \textbf{50.7} & 45.8 \\
\hline
$\ddagger$ Cylinder3D (baseline) & 71.7 & 97.1 & 69.6 & 72.0 & 55.8 & 62.4 & 76.2 & 77.8 & 46.7 & 91.2 & 69.8 & 76.2 & 40.9 & 92.6 & \textcolor{red}{70.2} & \textcolor{red}{86.7} & \textcolor{red}{73.8} & 71.6 & 65.2 & 66.3 \\
\textbf{Cylinder3D+SalLiDAR-I} & \textcolor{red}{\textbf{72.4}} & \textbf{97.2} & \textcolor{red}{\textbf{70.0}} & \textcolor{red}{\textbf{73.1}} & \textbf{59.7} & \textcolor{red}{\textbf{63.0}} & \textcolor{red}{\textbf{77.7}} & \textcolor{red}{\textbf{78.4}} & \textbf{50.4} & \textbf{91.3} & \textbf{70.5} & \textbf{76.3} & \textbf{41.3} & \textbf{92.6} & 69.9 & 86.5 & 73.4 & 70.8 & \textcolor{red}{\textbf{66.2}} & \textbf{66.4}  \\
\textbf{Cylinder3D+SalLiDAR-II} & \textbf{72.0} & \textbf{97.2} & 69.0 & \textbf{72.0} & \textbf{59.8} & \textbf{62.8} & \textbf{76.6} & 77.3 & \textbf{48.1} & \textbf{91.7} & \textcolor{red}{\textbf{70.9}} & \textbf{77.2} & \textcolor{red}{\textbf{41.6}} & 92.5 & 69.4 & 86.4 & 73.4 & 71.2 & \textbf{65.5} & 65.3  \\
\textbf{Cylinder3D+SalLiDAR-III} & \textbf{72.1} & \textbf{97.2} & 69.2 & \textbf{72.1} & \textcolor{red}{\textbf{60.6}} & \textcolor{red}{\textbf{63.0}} & \textbf{77.5} & 77.7 & \textbf{47.8} & \textcolor{red}{\textbf{91.8}} & \textbf{70.7} & \textbf{77.4} & \textbf{41.1} & \textbf{92.6} & 69.5 & 86.4 & 73.6 & 71.1 & \textbf{65.6} & 65.4  \\
\hline
\hline
\multicolumn{21}{l}{* PyTorch implementation of RandLA-Net~\cite{hu2020randla}, which is available at: {\url{https://github.com/tsunghan-wu/RandLA-Net-pytorch}}.} \\
\multicolumn{21}{l}{$\ddagger$ The results are obtained from the released version of Cylinder3D model \cite{zhu2021cylindrical} from the work in \cite{hou2022point}: {\url{https://github.com/cardwing/Codes-for-PVKD}}.} \\
\multicolumn{21}{l}{Best performance results are shown in \textcolor{red}{red} color (publications before July 2022). Improved performance results of proposed model against the baseline are shown in \textbf{bold}.} \\
\end{tabular}
\end{adjustbox}
\label{tab_semkitti_test}
\end{table*}

\subsubsection{Semantic Segmentation Results on SemanticKITTI Dataset}
\label{exp_results_semantic}

We report the LiDAR semantic segmentation performance on the test set of SemanticKITTI in Table~\ref{tab_semkitti_test}. Note that all the testing performance results of Table~\ref{tab_semkitti_test} are taken from the literature and the benchmark leaderboard\footnote{\url{http://www.semantic-kitti.org/tasks.html\#semseg}} of SemanticKITTI \cite{behley2019semantickitti} dataset.  By comparing with Table~\ref{tab_saliency_val} and Table~\ref{tab_semkitti_test}, we can find that the mIoU results on test sequences show the improved generalization ability in the larger set of evaluation samples. 
Compared to the baselines, all the models with SalLiDAR obtain better mIoU results. The proposed method also improves the segmentation performance on specific classes, since the combination of our predicted saliency distribution makes the model attentive to these categories, such as \textit{car}, \textit{truck}, and \textit{parking}. Furthermore, the Cylinder3D model with SalLiDAR achieves better segmentation results than the RandLA-Net with SalLiDAR. It shows that the semantic segmentation model with better saliency prediction could provide more attentive information or features to improve the performance of the model. Especially, these experimental results demonstrate that the performance of LiDAR semantic segmentation models can be improved by proposed saliency distribution integration and point-wise attention-guided loss. These comparison results validate the effectiveness of the pre-trained point cloud saliency models, although they are trained on the FordSaliency dataset with pseudo-annotations.

\begin{figure}[!ht]
    \centering
    \includegraphics[width = 0.5\textwidth]{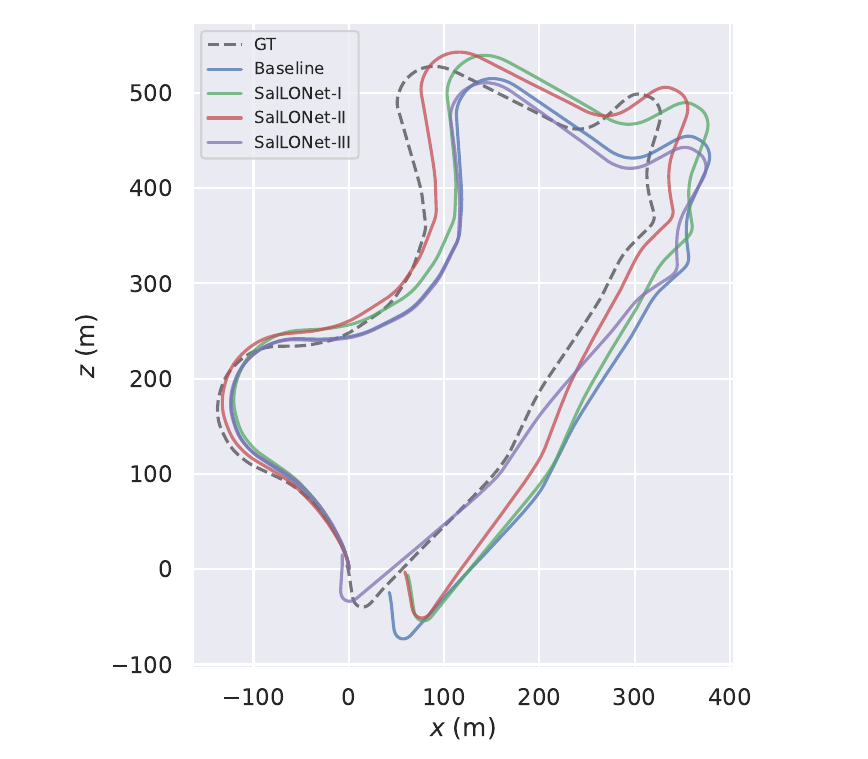} 
    \includegraphics[width = 0.5\textwidth]{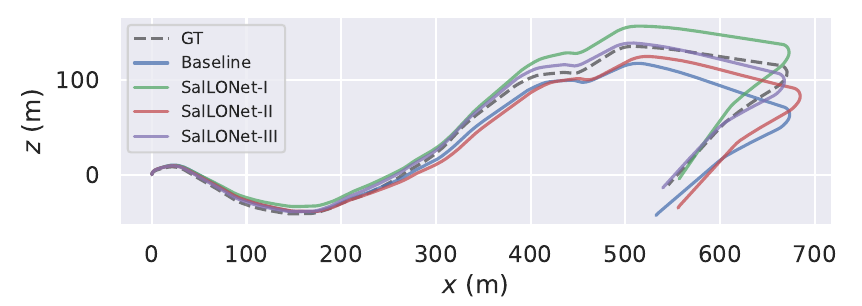} 
    \caption{Comparison trajectory results of proposed SalLONet odometry models against the baseline odometry model \cite{nubert2021self} on Sequence 09 (top) and Sequence 10 (bottom) of KITTI \cite{geiger2012we} odometry dataset.
    }
\label{fig:odom_traj_results}
\end{figure}

\begin{figure*}[!ht]
    \centering
    \includegraphics[width = 0.48\textwidth]{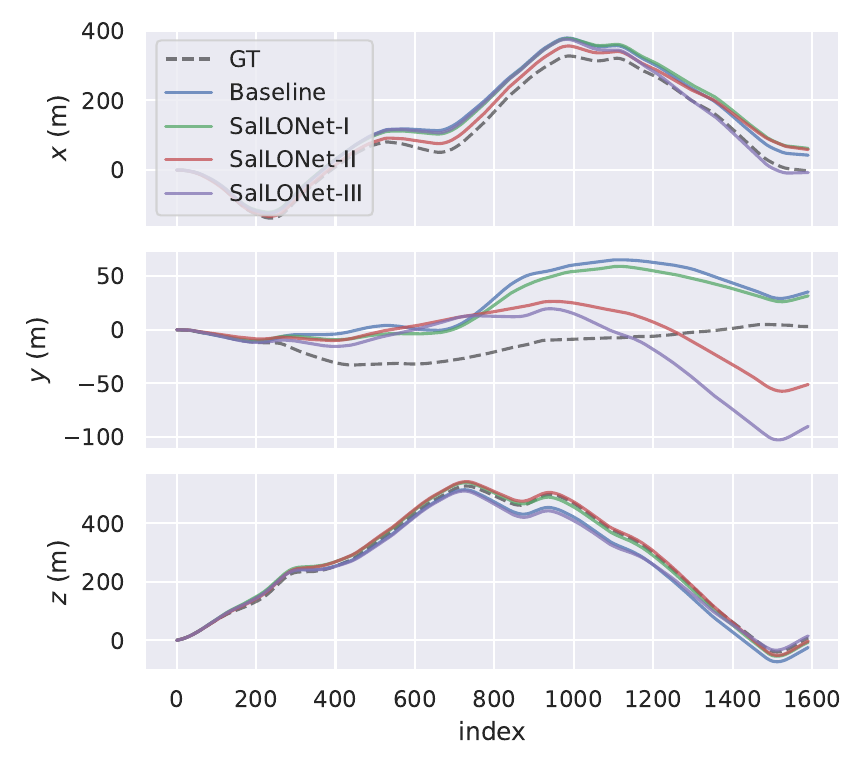} 
    \includegraphics[width = 0.48\textwidth]{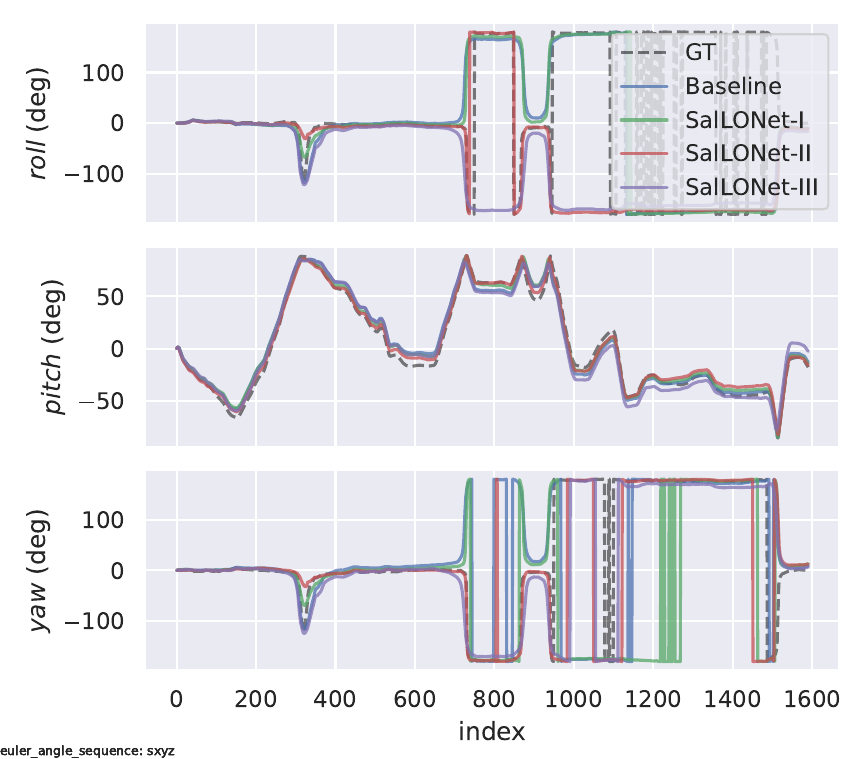} \\
    \vspace{-0.3cm}
    \includegraphics[width = 0.96\textwidth]{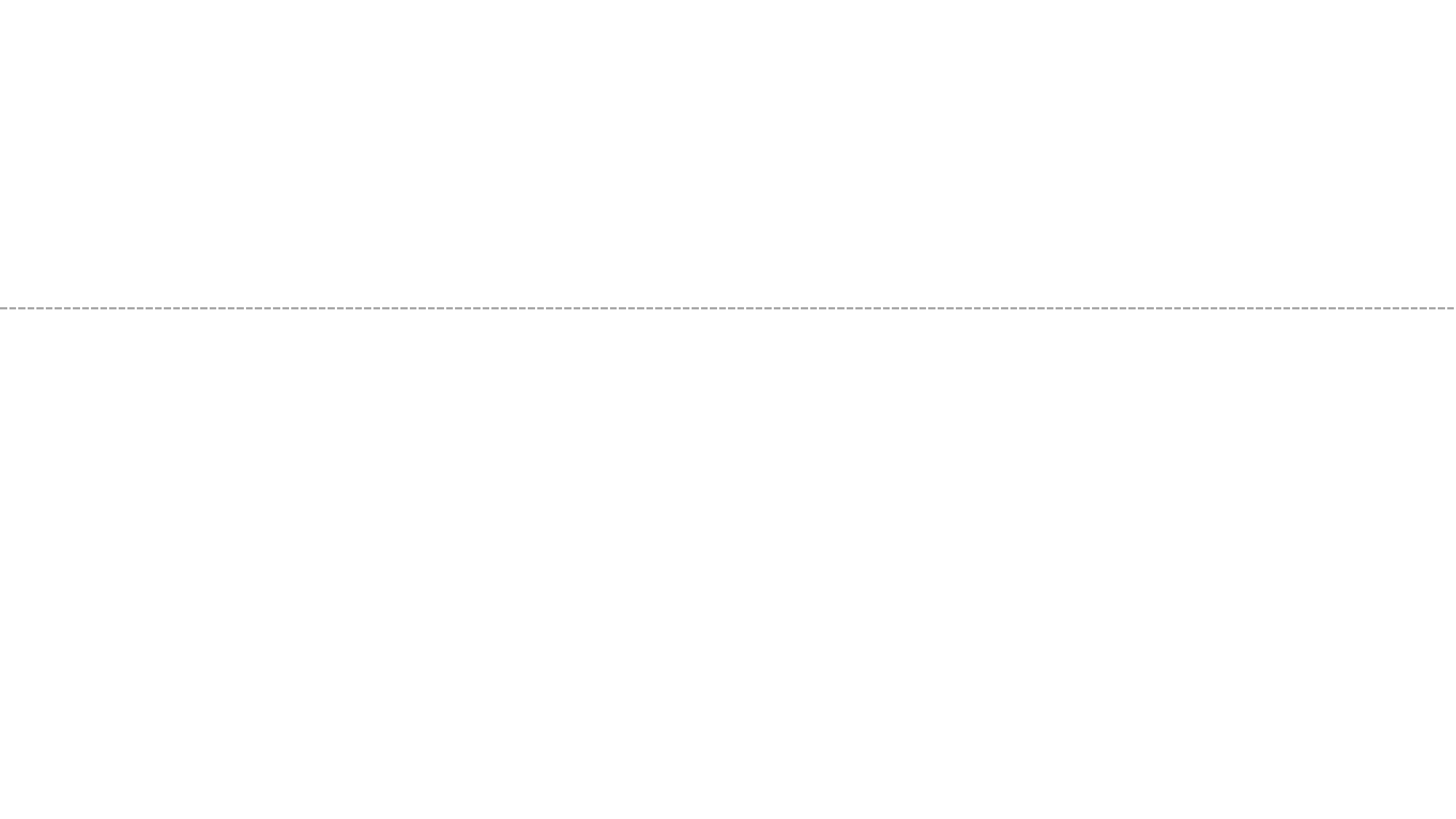} \\
    \includegraphics[width = 0.48\textwidth]{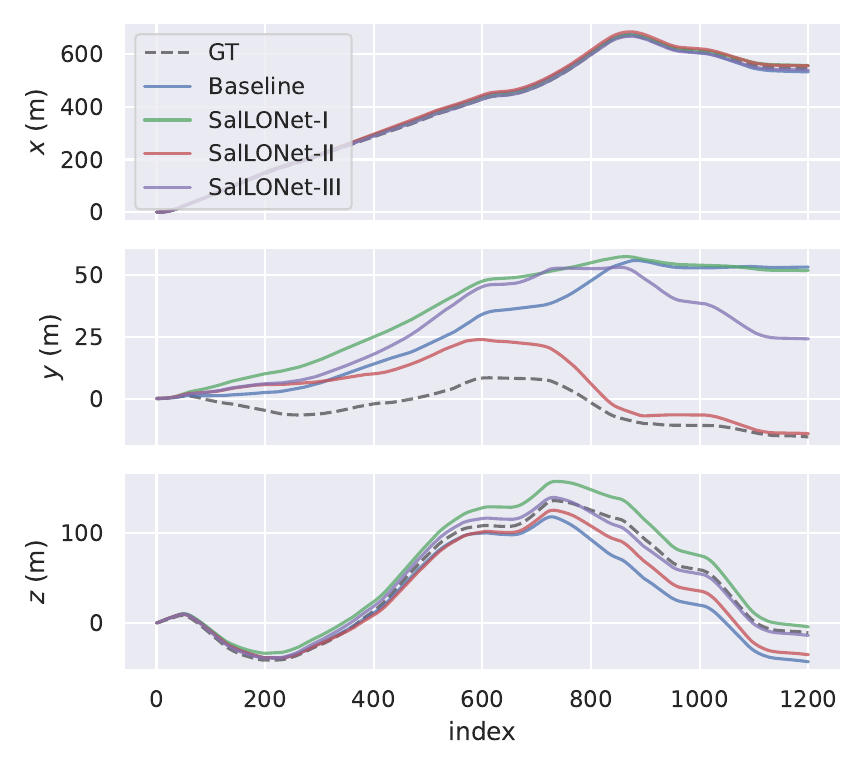} 
    \includegraphics[width = 0.48\textwidth]{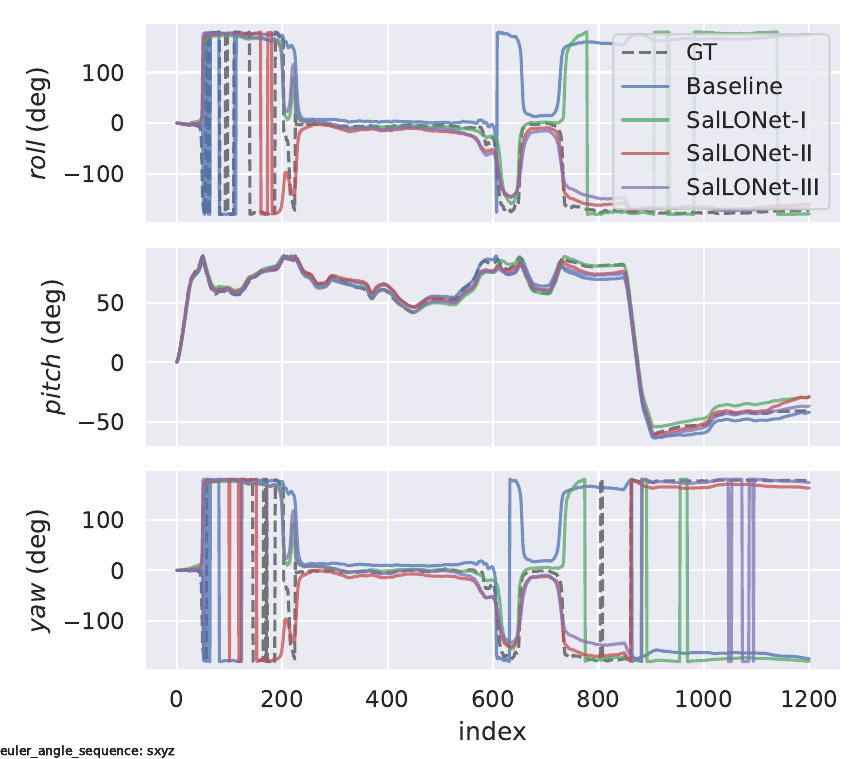}
    \caption{Detailed trajectory results of proposed SalLONet odometry models against the baseline odometry model \cite{nubert2021self} on Sequence 09 (top) and Sequence 10 (bottom) of KITTI \cite{geiger2012we} odometry dataset.
    }
\label{fig:odom_traj_results2}
\end{figure*}

\begin{table*}[t]
        \centering
        \caption{Comparison of translational ($[\%]$) and rotational ($[\frac{deg}{100m}]$) errors on validation set of KITTI \cite{geiger2012we} odometry dataset. The performance results are partially obtained from \cite{nubert2021self}. Compared with the baseline, improved results are shown in \textbf{bold}.
        }
        \resizebox{0.8\textwidth}{!}{%
        \begin{tabular}{c c c c c c c c} 
             \hline
             \hline
              ~ & Supervised/ & \multicolumn{2}{c}{Mean \texttt{09-10}} & \multicolumn{2}{c}{Sequence \texttt{09}} & \multicolumn{2}{c}{Sequence \texttt{10}} \\
              Model & Unsupervised & $t_{\text{rel}}$ $\downarrow$ & $r_{\text{rel}}$ $\downarrow$ & $t_{\text{rel}}$ $\downarrow$ & $r_{\text{rel}}$ $\downarrow$ & $t_{\text{rel}}$ $\downarrow$ & $r_{\text{rel}}$ $\downarrow$  \\ 
              \hline
              DeepLO \cite{cho2019deeplo} & Supervised & 4.945 & 1.890 & 4.870 & 1.950 & 5.020 & 1.830 \\
              Velas \emph{et al.} \cite{velas2018cnn} & Supervised & 4.105 & NA & 4.940 & NA & 3.270 & NA \\
              \hline
              DeLORA \cite{nubert2021self} & Unsupervised & 6.245 & 2.575 & 6.050 & 2.150 & 6.440 & 3.000 \\
              Zhu \emph{et al.} \cite{zhu2018robustness} & Unsupervised & 7.745 & 3.405 & 8.840 & 2.920 & 6.650 & 3.890 \\
              SfMLearner \cite{zhou2017unsupervised} & Unsupervised & 16.550 & 3.255 & 18.770 & 3.210 & 14.330 & 3.300 \\
              UnDeepVO \cite{li2018undeepvo} & Unsupervised & 8.820 & 4.130 & 7.010 & 3.610 & 10.630 & 4.650 \\
              \hline
              *DeLORA (Baseline) & Unsupervised & 6.420 & 2.718 & 6.764 & 2.828 & 6.076 & 2.608 \\
              \textbf{SalLONet-I} & Unsupervised & \textbf{6.320} & \textbf{2.426} & 7.740 & \textbf{2.595} & \textbf{4.900} & \textbf{2.257} \\
              \textbf{SalLONet-II} & Unsupervised & 6.438 & \textbf{2.574} & 7.424 & \textbf{2.515} & \textbf{5.451} & 2.632 \\
              \textbf{SalLONet-III} & Unsupervised & \textbf{5.254} & \textbf{2.339} & \textbf{5.567} & \textbf{2.088} & \textbf{4.940} & \textbf{2.590} \\
             \hline
             \hline   
            \multicolumn{8}{l}{\small *The results of the baseline model are obtained by re-training the model from scratch.}\\
        \end{tabular} }
        \label{table:delora_sallidar_eval}
    \vspace{-0.2cm}
\end{table*}

\subsubsection{Odometry Results on KITTI Dataset}
\label{exp_results_odometry}
In Figure \ref{fig:odom_traj_results} and Figure \ref{fig:odom_traj_results2}, we show the experimental trajectory results of the proposed SalLONet models on Sequence 09-10 of KITTI \cite{geiger2012we} odometry dataset. We can observe that the SalLONet models with saliency and semantic information predict better trajectory results than the baseline model by comparing with the ground truth. In Table \ref{table:delora_sallidar_eval}, we present the quantitative results of proposed approaches and six existing odometry methods. The DeepLO \cite{cho2019deeplo} and Velas \emph{et al.} \cite{velas2018cnn} are supervised LiDAR odometry models. In other words, the ground-truth poses of the training set are used to train these supervised odometry models. The DeLORA \cite{nubert2021self} is an unsupervised LiDAR odometry model. This means that the unsupervised DeLORA \cite{nubert2021self} does not require labels to train the model. By following the study \cite{nubert2021self}, there are also three unsupervised visual odometry estimation methods \cite{zhu2018robustness, zhou2017unsupervised, li2018undeepvo} for comparison. As shown in Table \ref{table:delora_sallidar_eval}, the three proposed SalLONet models improve the performance of the baseline model, as evidenced by lower translational and rotational errors in Sequence 09-10 of KITTI \cite{geiger2012we} odometry dataset. Among the unsupervised methods, the SalLONet-III achieves the best results with the lowest errors on both validation sequences. In particular, its translational error on Sequence 10 ($t_{\text{rel}}$=4.940) even outperforms the supervised DeepLO \cite{cho2019deeplo} method ($t_{\text{rel}}$=5.020).
In a word, these experimental results show that the saliency and semantic information are effective for improving the odometry estimation tasks, which implicitly indicates the effectiveness of image-to-LiDAR saliency knowledge transfer.

\begin{table*}[t]
        \centering
        \caption{Comparison of translational ($[\%]$) and rotational ($[\frac{deg}{100m}]$) errors on validation set of KITTI \cite{geiger2012we} odometry dataset. By comparing with the baseline model, improved results are shown in \textbf{bold}.
        }
        \resizebox{0.8\textwidth}{!}{%
        \begin{tabular}{c c c c c c c} 
             \hline
             \hline
              ~ & \multicolumn{2}{c}{Mean \texttt{09-10}} & \multicolumn{2}{c}{Sequence \texttt{09}} & \multicolumn{2}{c}{Sequence \texttt{10}} \\
              Model & $t_{\text{rel}}$ $\downarrow$ & $r_{\text{rel}}$ $\downarrow$ & $t_{\text{rel}}$ $\downarrow$ & $r_{\text{rel}}$ $\downarrow$ & $t_{\text{rel}}$ $\downarrow$ & $r_{\text{rel}}$ $\downarrow$  \\ 
              \hline
              *DeLORA (Baseline) & 6.420 & 2.718 & 6.764 & 2.828 & 6.076 & 2.608 \\
              SalLONet-III w/ saliency-only & \textbf{5.705} & \textbf{2.584} & 7.157 & \textbf{2.559} & \textbf{4.253} & {2.609} \\
              SalLONet-III w/ semantic-only & 7.605 & 3.000 & 8.400 & 3.064 & 6.810 & 2.937 \\
              \textbf{SalLONet-III} & \textbf{5.254} & \textbf{2.339} & \textbf{5.567} & \textbf{2.088} & \textbf{4.940} & \textbf{2.590} \\
             \hline
             \hline   
            \multicolumn{7}{l}{\small *The results of the baseline model are obtained by re-training the model from scratch.}\\
        \end{tabular} }
        \label{table:sallonet_ablation}
    \vspace{-0.2cm}
\end{table*}

\subsection{Ablation Studies}
\label{exp_results_ablation}
We investigate the effectiveness of saliency and semantics for LiDAR odometry estimation. From Table \ref{table:delora_sallidar_eval}, we observe that the SalLONet-III model achieves better results by leveraging both semantic and saliency cues. Thus, we conduct the ablation study based on SalLONet-III to verify the influences of saliency and semantic maps for LiDAR odometry estimation. In the proposed SalLONet-III method, we leverage saliency and semantic predictions for LiDAR odometry estimation simultaneously. We first validate the model with saliency information-only integration. We also train the model with semantic information only. The performance results of the ablation study are shown in Table \ref{table:sallonet_ablation}. Experimental evaluation shows that the SalLONet model with both saliency and semantic information achieves superior performance on KITTI \cite{geiger2012we} odometry dataset. In addition, the model of SalLONet with saliency only outperforms better results against the baseline model, which demonstrates that saliency information is effective for improving LiDAR odometry estimation. The model of SalLONet with semantics only obtains competitive results by comparing it with the baseline model. However, the SalLONet model benefits from saliency and semantic information, thus getting the lowest translational and rotational errors on Sequence 09-10 of the KITTI \cite{geiger2012we} odometry dataset.

\section{Conclusion}
\label{sec_Conclusion}
This article has presented the research works on establishing LiDAR-based saliency detection models with image-to-LiDAR transfer learning for improving the performance of 3D point cloud understanding tasks.
We propose a \textbf{Sal}iency-guided \textbf{L}iDAR \textbf{O}dometry \textbf{Net}work (\textbf{SalLONet}) by combining saliency and semantic information of point clouds. First, the saliency and semantic maps generated by the proposed two-stream semantic model are fed into the odometry module as the feature representation of the input consecutive point clouds. Second, the saliency and semantic predictions are applied to odometry loss. To alleviate the effect of dynamic points for pose regression, we binarize the semantic prediction to dynamic and static points based on the semantic class. Then the binarized semantics are utilized to filter the dynamic points by point-wise multiplication for loss weighting. To further encourage the odometry module to learn discriminative features, the saliency map is leveraged to increase the loss weights of salient static points for matching two-point clouds. Extensive experimental results on KITTI \cite{geiger2012we} odometry dataset have demonstrated outstanding performance of the proposed odometry model with saliency and semantic information, which considers the influences of dynamic and static salient points for pose estimation simultaneously.

\section*{ACKNOWLEDGMENT}
\label{sec:ACKNOWLEDGMENT}
This paper is in part based on the results obtained from a project commissioned by the New Energy and Industrial Technology Development Organization (NEDO), Japan. This work was supported by JST SPRING, Grant Number JPMJSP2124. Computational resource of \emph{AI Bridging Cloud Infrastructure (ABCI)\footnote{https://abci.ai/}} provided by the National Institute of Advanced Industrial Science and Technology (AIST) was used for training and testing the models during our experiments.

\bibliography{References}

\end{document}